\documentclass[lettersize,journal]{IEEEtran}

\usepackage[colorlinks,urlcolor=blue,linkcolor=blue,citecolor=blue]{hyperref}

\usepackage{microtype}
\usepackage{graphicx}
\usepackage{multirow}
\usepackage{makecell}
\usepackage{booktabs} 
\usepackage{paralist}
\usepackage{cite}
\usepackage{url}
\usepackage{hyperref}
\usepackage{xspace}

\usepackage{enumerate}
\usepackage[utf8]{inputenc} 
\usepackage[T1]{fontenc}    
\usepackage{url}            
\usepackage{booktabs}       
\usepackage{amsfonts}       
\usepackage{amsmath}       
\usepackage{nicefrac}       
\usepackage{microtype}      
\usepackage{algorithmicx}
\usepackage{algorithm}
\usepackage{algpseudocode}

\usepackage{mwe}    
\usepackage{float}
\usepackage{lineno}
\usepackage{caption}
\usepackage[skip=0cm,list=true]{subcaption}
\usepackage[dvipsnames]{xcolor}
\definecolor{codegreen}{rgb}{0,0.6,0}
\definecolor{codegray}{rgb}{0.5,0.5,0.5}
\definecolor{codepurple}{rgb}{0.58,0,0.82}
\definecolor{backcolour}{rgb}{0.95,0.95,0.92}

\newcommand{\smartparagraph}[1]{\noindent{\bf #1}\ }

\usepackage[capitalise]{cleveref}
\usepackage{paralist}

\begin{document}

\title{Leveraging The Edge-to-Cloud Continuum for Scalable Machine Learning on Decentralized Data}

\author{Ahmed M. Abdelmoniem$^*$,~\IEEEmembership{Member,~IEEE}
\IEEEcompsocitemizethanks{
\IEEEcompsocthanksitem $^*$Corresponding Author, E-mail: ahmed.sayed@qmul.ac.uk
\IEEEcompsocthanksitem Ahmed is an Assistant Professor at Queen Mary University of London, UK. Webpage: \url{http://eecs.qmul.ac.uk/\~ahmed}
}
}


\maketitle

\begin{abstract}
With mobile, IoT and sensor devices becoming pervasive in our life and recent advances in Edge Computational Intelligence (e.g., Edge AI/ML), it became evident that the traditional methods for training AI/ML models are becoming obsolete, especially with the growing concerns over privacy and security. This work tries to highlight the key challenges that prohibit Edge AI/ML from seeing wide-range adoption in different sectors, especially for large-scale scenarios. Therefore, we focus on the main challenges acting as adoption barriers for the existing methods and propose a design, with a drastic shift from the current ill-suited approaches, that leverages the Edge-to-Cloud continuum. The new design is envisioned to be model-centric in which the trained models are treated as a commodity driving the exchange dynamics of collaborative learning in decentralized settings. It is expected that this design will provide a decentralized framework for efficient collaborative learning at scale.
\end{abstract}







\section{Introduction}
\label{sec:intro}
Due to the proliferation of mobile, IoT, sensors, and edge devices in our daily life activities which produce tremendous amounts of data, AI/ML-based analytics, which can process these larger volumes of data, became an integral part of user products and applications as well as the key revenue source for most organizations~\cite{kairouz2019advances,Letaief2019roadmap6G}. These data-driven analytics rely on data which is produced on edge devices for training the AI/ML models~\cite{kairouz2019advances,Letaief2019roadmap6G}. Traditionally, the data were collected and stored in the cloud where the analytics and intelligence are, however, recently \textbf{Edge AI} paradigm was coined to move the analytics and intelligence where the data is. This is also driven by the recent security and privacy concerns over collecting or storing private user data~\cite{FL_IoTM}.

Within the Edge AI paradigm, several methods have evolved to achieve the learning task on the decentralized data such as:
\begin{inparaenum} 
\item Federated Learning (FL)~\cite{mcmahan2017}: which runs training locally on the devices and the server aggregates the model updates and coordinates the training rounds;
\item Split Learning (SL)~\cite{splitlearning} which splits the training task between the device and the cloud server (or edge server);
\item Decentralized Learning (DL)~\cite{cartel-Daga2019}: which leverages peer-to-peer coordination for the model exchange among the edge devices.
\end{inparaenum} 
These approaches are made possible by harnessing the advances in the AI/ML accelerators embedded in edge devices~\cite{kairouz2019advances} and the high-throughput and low-latency 5G/6G technologies~\cite{Letaief2019roadmap6G}. However, several challenges manifest themselves impacting the efficiency of these methods and making them ill-suited for decentralized Edge AI at scale. The highly heterogeneous devices, configurations, and environment and the strict synchronization requirements are among the key challenges. This not only results in models with low qualities and long training times but also hinders the existing approaches from scaling with a large number of learners ~\cite{kairouz2019advances,Ahmed-EuroMLSys-22,Ahmed-REFL-2023,FL_IoTM}. It is evident that low-quality models can be costly to many businesses and organizations. For instance, recently the real-estate Zillow group wrote down \$304M in inventory due to the low accuracy of their Zestimate AI algorithm)~\cite{zillow}.

It is clear that it modern data-driven intelligence needs systems that can produce accurate and timely models~\cite{kairouz2019advances,FL_IoTM}. In this work, we shed light on the key challenges in existing decentralized methods and try to propose a novel architectural design for scalable decentralized learning. To this end, we aim to leverage the Edge-to-Cloud continuum~\cite{edgecloud-2020} and transform the learning task into a collaborative knowledge transfer system which facilities for the learning parties to share (or trade) the trained models based on mutual benefits or needs. In this view, the trained AI/ML models are treated as a commodity that can be exchanged between learning entities to meet global or personal objectives. This view is also similar to the majority of online delivery services such as Uber (passengers) or Deliveroo (food). For example, Uber's task is merely connecting the passengers wanting to be transported from point A to point B with the appropriate drivers. Similarly, in this work, the proposed idea facilitates the exchange and delivery of the trained models among the learning entities. In particular, we seize an opportunity to decouple the training task of a common model from the sharing task of the trained models. This would facilitate the exchange of models among learning parties and significantly improve AI/ML-based analytics of data generated by various connected devices and sensors that are now an integral part of our daily-life applications. 

\section{Background}

\begin{figure*}
    \centering
    \includegraphics[width=1\linewidth]{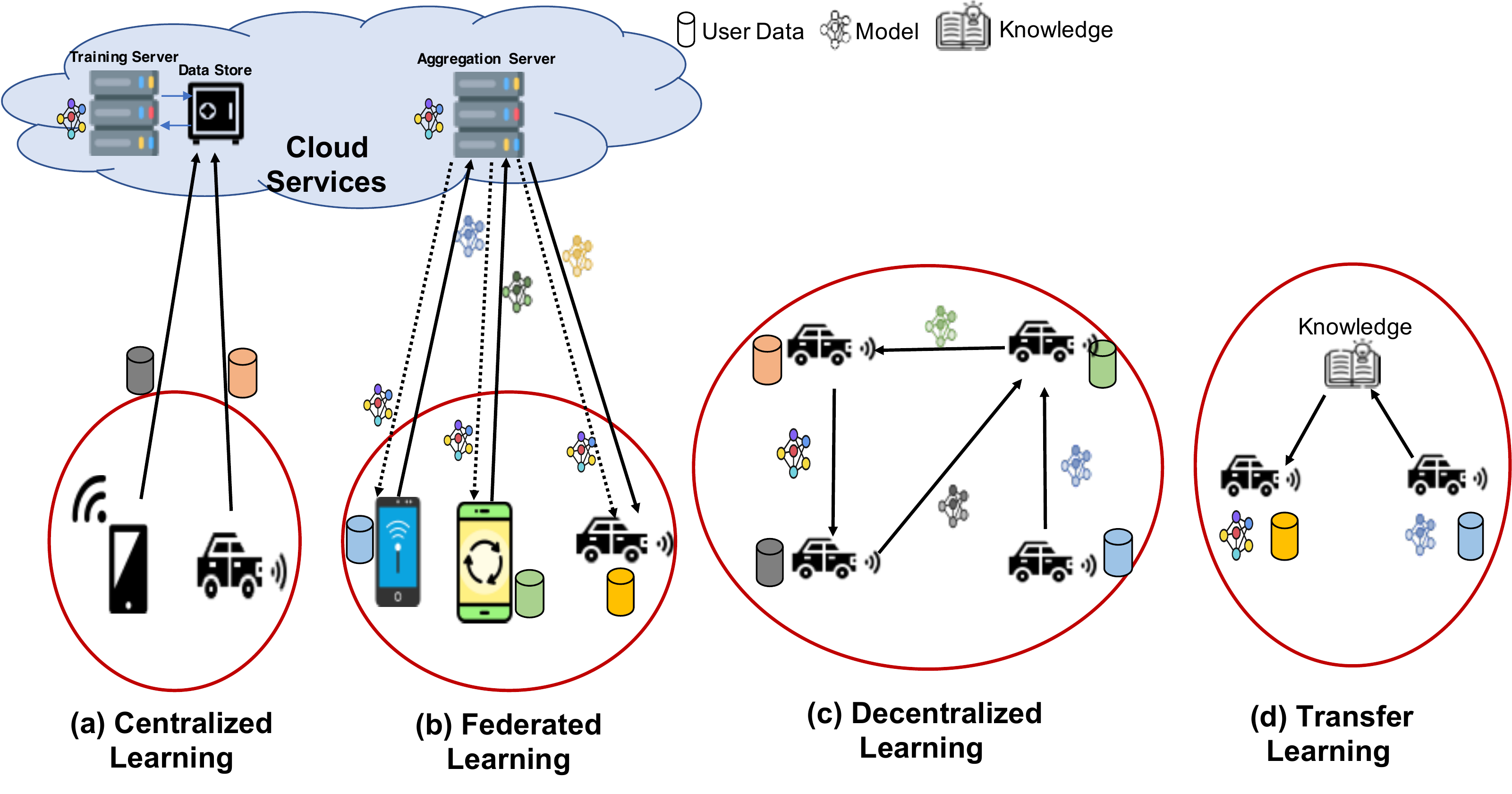}
    \caption{Comparison of the existing traditional learning paradigms. (a) centralized learning - involving data collection and then training. (b) federated learning - involving clients training local models and aggregating them to a global model via a server. (c) decentralized learning - clients train collaboratively on a common model via peer-to-peer exchange and coordination. (d) transfer learning - the knowledge within models is used as the basis for training other models via means of distillation.}
    \label{fig:traditonal}
\end{figure*}

We present the various methodologies traditionally used for training AI/ML models on big data generated from large-scale IoT, sensor and mobile devices. \cref{fig:traditonal} contrasts the common four approaches for learning of largely-decentralized data.

\smartparagraph{(a) Centralized Learning (CL) - \cref{fig:traditonal}(a): } typically, these devices continuously stream the generated data into the cloud applications to be stored for later processing and analysis. These data are analyzed to extract certain features to help train AI/ML models on the cloud. These models are trained on high-performance servers residing in the data centres of the cloud. Google Cloud, Azure, and AWS are the most common providers for ML-as-a-service where models can be trained on large amounts of data at scale. Unfortunately, Centralized learning (CL) is becoming obsolete due to the collection of user data imposing security and privacy risks and expensive communication costs for the data transfer~\cite{Bonawitz19,cartel-Daga2019,kairouz2019advances,SecureFTL}. 

\smartparagraph{b) Federated Learning (FL) - \cref{fig:traditonal}(b):} In the FL paradigm, the clients are users with end devices (e.g., mobile, IoT, or sensors) using apps generating the data. The data is securely stored on the device and should not leave it. So, the clients train local models and engage with the FL server to create a global model. The server assists with the  aggregation of the clients' local models~\cite{mcmahan2017,Bonawitz19,FL_IoTM}. 
The training process happens in a sequence of rounds controlled by the server until a certain objective is met (e.g., target accuracy). In each round, a few clients are sampled to update the model and a new model is produced. Due to the server's role, FL faces challenges of synchronization, reliability, and expensive communication~\cite{kairouz2019advances}. When a training round starts, the available clients start to login-and express their interest to participate with the server. A client is typically available if it is connected to WiFi and a power source and is idle~\cite{Bonawitz19}. The server then executes a selection algorithm to choose a subset of the large client population. Then, the server sends to the selected clients the FL task (i.e., the ML model and configurations of the hyper-parameters). The clients perform the same number of local optimization steps as set by the task. Then, the clients upload their updated local models to the server. Then, the FL server aggregates the uploaded client models to produce a new global model which is check-pointed to the local storage of the server~\cite{Bonawitz19}.

\smartparagraph{(d) Decentralized Learning (DL) -  \cref{fig:traditonal}(c):} 
DL is a different method for developing generic models on decentralized data in environments with edge devices \cite{cartel-Daga2019}. Peer-to-peer communication is used by the learners in decentralized learning (DL) to coordinate among themselves while training a model that is tailored to their shared tasks. As a result, device groups can train a shared model while preserving the data of individual devices. The devices must always be present to iterate over the training process in a lock-step manner, and stragglers slow down the training because there is no central coordination. Scalability and efficiency are hampered as a result, as devices cannot train independently of slow learners \cite{cartel-Daga2019, kairouz2019advances}.

\smartparagraph{(e) Transfer Learning (TL) - \cref{fig:traditonal}(d):} 
TL is a machine learning technique that transfers information from one domain to another. TL relies on previously learned information that ML algorithms can re-purpose, according to \cite{TLsurvey}. In addition, TL encourages the use of IID data in situations where there are insufficient training data \cite{TLsurvey}. IID data availability is no longer reliant on their availability. However, it isn't easy to find and choose which model to use for the transfer as there is a large pool of models to choose from~\cite{CollabEqualib}. It is also challenging to perform the distillation of knowledge within the trained models efficiently~\cite{TLsurvey}.

\section{Challenges and Motivation}
We have covered the traditional learning paradigms and highlighted their key differences. We now focus on the critical challenges of these paradigms that result in inefficient learning and low scalability.  

One of the detrimental challenges that the traditional paradigms face, which hinders efficiency and scalability, is the heterogeneity of the users and learning environment~\cite{Bonawitz19,kairouz2019advances,FL_IoTM}. When training on decentralized parties, the common types of heterogeneity can be one or the combination of: 
\begin{itemize}
\item \textbf{data}: non-uniform data points of the learners in number, type and distribution;
\item \textbf{device}: diverse set of hardware and network settings;
\item \textbf{behavioural}: dynamic participation availability which is driven by learners' behaviour.
\end{itemize}
In \cref{sec:eval}, we present experimental results covering over 1,500 different scenarios of training five different models on five different federated datasets decentralized over large populations of thousands of users. As demonstrated by the results, heterogeneity, regardless of type, can significantly impact the model quality (as well as fairness) and cause model divergence in the worst case. Since heterogeneity is endemic to decentralized learning, mitigating its impact on the performance of decentralized learning methods is desirable. 

We identify a pressing need to manage, process and analyze the increasing amount of decentralized data generated by collaborative parties to support the growth of machine learning and data-driven applications ~\cite{Bonawitz19,kairouz2019advances,Letaief2019roadmap6G,cartel-Daga2019}. We seize an opportunity from the proliferation of connected devices and edge computing which create a wealth of data at the network's edge. However, there is a lack of frameworks that can do so efficiently, at a low cost, and with fast training time while observing privacy requirements. Without questions, any centralized approach that collects user data becomes infeasible. Nowadays, it is widely accepted that data is processed locally on edge devices and models will be trained via collaborative approaches on decentralized data. This will be the key driver for high-quality, personalized, and intelligent services for a customized user experience, including personalized healthcare, gaming, and recommendation services~\cite{Letaief2019roadmap6G,kairouz2019advances,splitlearning}. These forms of decentralized learning not only need to preserve data privacy but also need to consider the efficient use of intelligence and computational capabilities available  on the edge.

Unfortunately, the current decentralized paradigms such as federated or decentralized learning can not scale to address the increasing demand and adoption of decentralization for services and applications~\cite{kairouz2019advances}. Moreover, practical deployment aspects like heterogeneity, communication costs, coordination, synchronization and availability to participate in training are all factors that either cause hinder the scalability, slow down the training process and/or affect the overall model quality (generalization performance)~\cite{kairouz2019advances,Ahmed-EuroMLSys-22,Ahmed-IoTJ-2023}. As a side effect, when training is not sufficiently fast, the models can not adapt to localized shifts in data trends~\cite{kairouz2019advances}.

The key question that forms the basis of this work: \textit{\bf What would be the best architectural design for collaborative learning over decentralized data?} 
We position this work as a call for architectural designs that simultaneously can:
\begin{enumerate}
 \item efficiently decentralize training over learning parties; 
 \item securely store a representation of the models in vaults; 
 \item develop a scalable model discovery and exchange methods.
\end{enumerate}
Next, we present a potential design of a decentralized architecture that tries to capture the essence of the above objectives.

\section{Design}
\label{sec:design}

\begin{figure}[!t]
    \centering
    \includegraphics[width=1\linewidth]{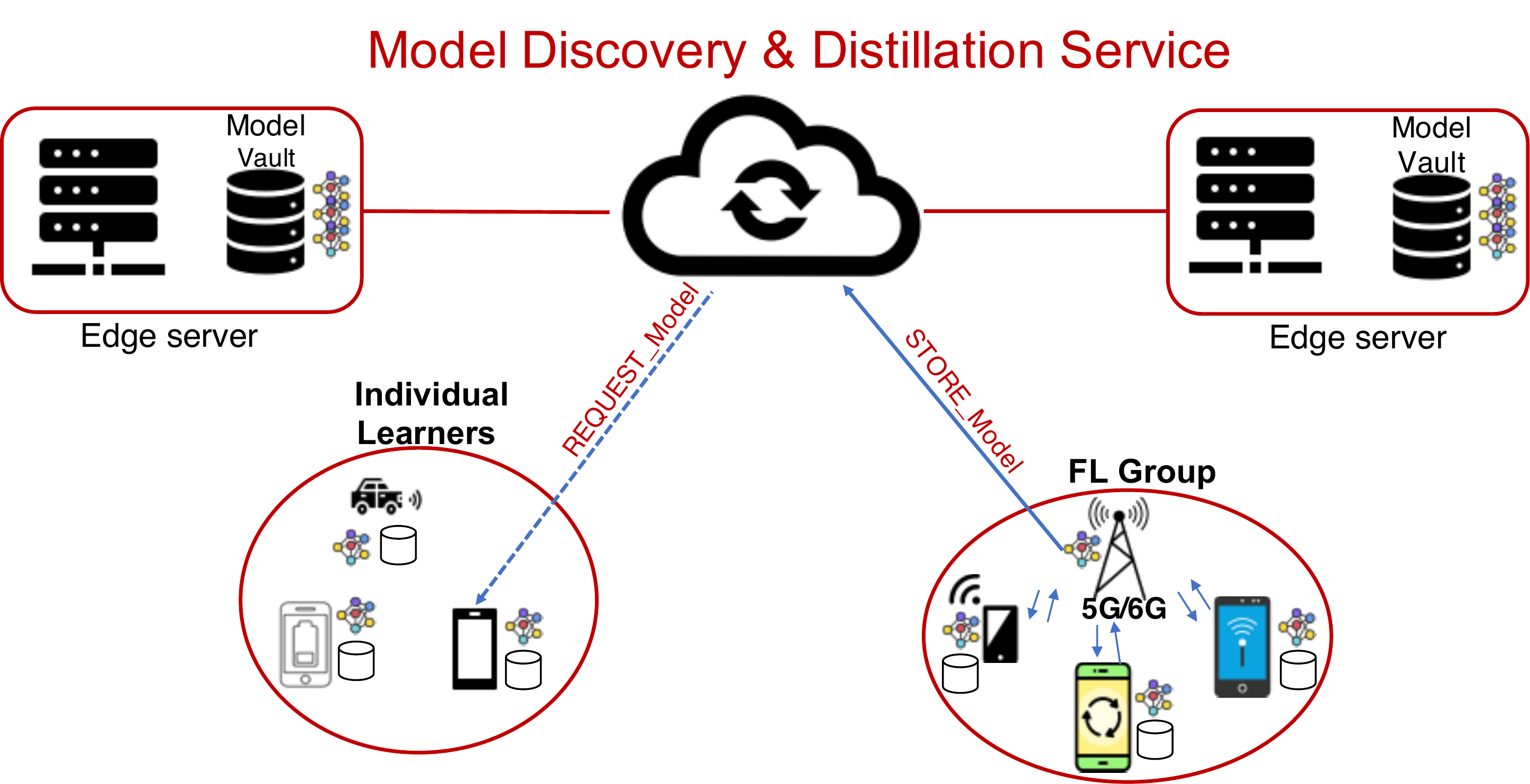}
    \caption{The proposed architectural design involves decentralized learning parties, secure model vaults to store the models hosted by edge servers and discovery service for model exchange hosted in the cloud.}
    \label{fig:system}
\end{figure}

We present the proposed architecture in \cref{fig:system}. The system is client-driven where learners create an initial model with their local datasets and then asynchronously seek models in the network to boost the model quality via means of distillation. The system discovers the model for the requester without involving other learners which helps mitigate any influence of the clients' heterogeneity. The proposed architecture may also introduce incentive mechanisms (e.g., based on monetary income or mutual interest~\cite{CollabEqualib}) to enable sharing of high-quality models in the network. The advantage of the proposed architecture is that it has no single point of control and failure (like FL), no explicit synchronization among devices (like DL) and no expensive movement of private data (like CL).

In the proposed architecture, as shown in \cref{fig:system}, the data owners train an initial model individually or in groups (e.g., via federated learning). Then, the learner(s) request to store the model in private and secure model stores (or vaults). The system will evaluate the model either on a public dataset by the service or via requesting testing parties to obtain the quality metrics of the model. The learning parties whenever they require improvements on their model, they send a request for a trained model to the discovery service specifying certain qualities of the requested model (e.g., a classifier needs to improve its accuracy to reach at least 90\% of accuracy for class D). The key innovation in this architecture is the design of the discovery service which requires novel discovery algorithms and protocols for finding the best models in the network fulfilling the requested qualities.\footnote{The exploration of scalable discovery algorithms is part of our future work}. Upon completion of discovery, the requester obtains the model and applies transfer learning (e.g., model distillation)~\cite{transferlearning,knowledgedistill-lan2018} to integrate the  new model into its own model and enhance its quality (e.g., classification accuracy for class D).

\section{Evaluation}
\label{sec:eval}

\begin{figure}[!t]
    \centering
\includegraphics[width=1\columnwidth]{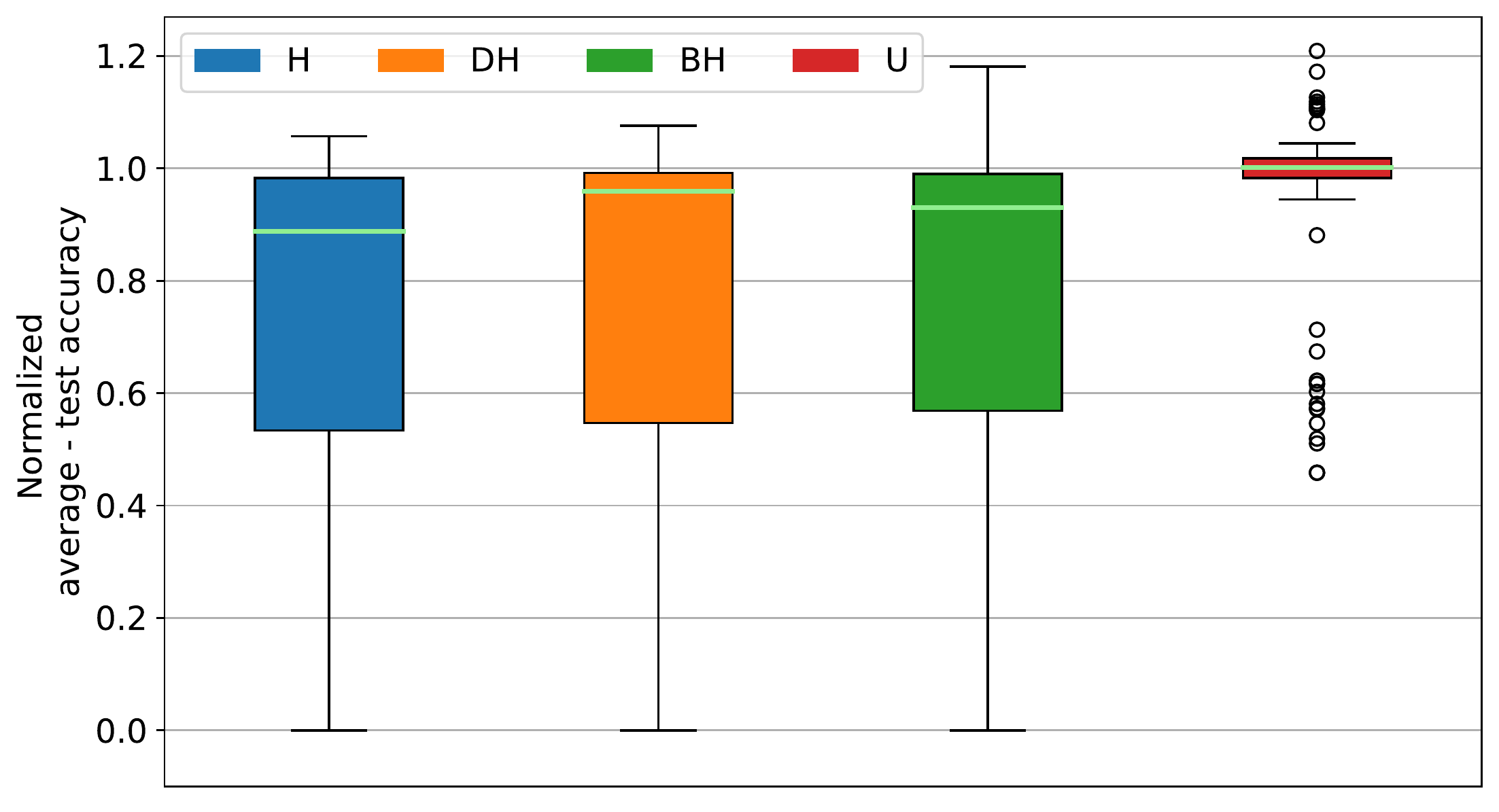}
    \caption{Impact of heterogeneity on the global model quality.}
    \label{fig:motivation}
\end{figure}

We use FLASH framework~\cite{yang2020heterogeneityaware} covering FL benchmarks of 6 different datasets involving tasks such as logistic regression, Computer Vision, and Natural Language Processing. The datasets are Non-IID reflecting the real-world data distribution among independent data owners. We extensively run experiments covering over 1,500 different heterogeneity, configurations and hyper-parameters settings. 

\smartparagraph{Experimental Setup:}
We partition the training and testing datasets so that each client owns a partition of each dataset as done in prior works. During a training round, a selected client uses samples from its training partition. We instrument the framework with information logging and configurations of nearly 1500 different cases (for more details c.f.~\cite{Ahmed-EuroMLSys-22,}).

\subsection{Heterogeneity Impact} 
As discussed previously, one of the key challenges in FL use cases is the heterogeneity of the environment~\cite{Bonawitz19,kairouz2019advances}.  \cref{fig:motivation} shows the average test accuracy (normalized to the maximum baseline accuracy for each of the 6 benchmarks) and contrasts the following cases :
\begin{inparaenum}[i)]
\item \textbf{\textcolor{red}{U}}: Uniform or homogeneous case where all clients are of the same hardware and network configuration and are always available. 
\item \textbf{\textcolor{ForestGreen}{BH}}: A behaviour-heterogeneity case where the devices have variable availability patterns based on real-world trace;
\item \textbf{\textcolor{orange}{DH}}: A device-heterogeneity case where the devices follow  and behaviour heterogeneity;
\item \textbf{\textcolor{blue}{H}}: A mixed-heterogeneity case with both device and behaviour heterogeneity
\end{inparaenum}
As Figure \ref{fig:motivation} shows, all the heterogeneity cases (\textcolor{ForestGreen}{\bf BH}, \textcolor{orange}{\bf DH}, and \textcolor{blue}{\bf H}) cases significantly impact the average test accuracy of the global model compared to the homogeneous case (\textcolor{red}{\bf U}). The results show that the median accuracy (the light green line) is lowered by up to 12\% and the variance is severely high in the heterogeneous cases. What makes matters worse is that the heterogeneity can cause the training to not converge. This reinforces the motivation of this work, that heterogeneity is endemic to decentralized learning settings and so the training at scale should cope with heterogeneity to produce high-quality models.

\begin{figure}[!h]
    \centering
\includegraphics[width=1\columnwidth]{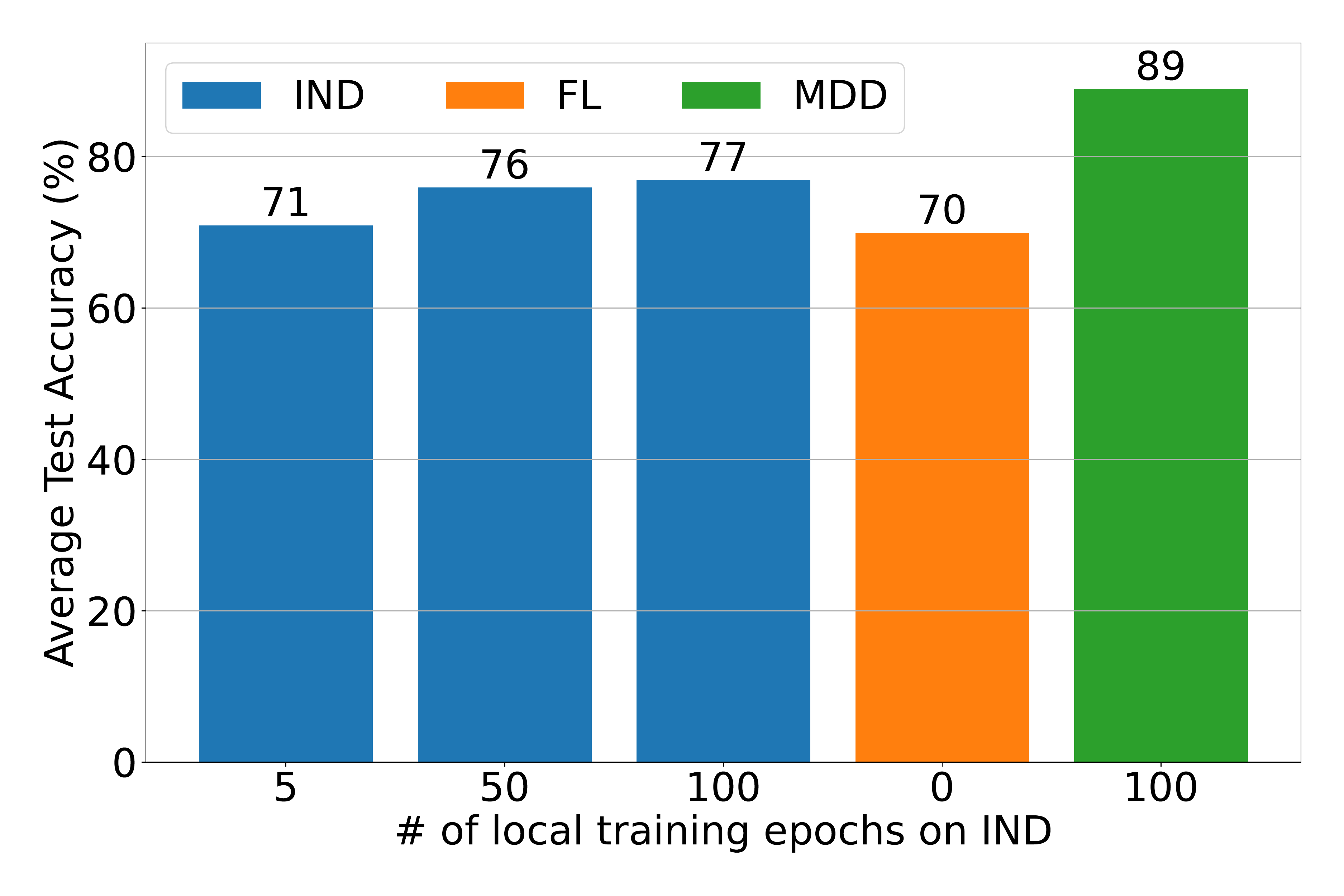}
    \caption{Performance comparison in the LR-Synthetic scenario.}
    \label{fig:results-synthetic}
\end{figure}

\subsection{Efficiency of The Proposed Approach} 
The experiments involve a small group of 10 independent parties individually training the model on their local data (\textcolor{blue}{\bf IND}) for a variable number of epochs and another large group consisting of the remaining clients training a global model via federated learning over 50 global rounds (\textcolor{ForestGreen}{\bf FL}). We contrast the model quality of these two groups with the case when the proposed architecture is used (\textcolor{red}{\bf MDD}). The 10 individual learners will request from the discovery service a global model trained by the FL group and then each of the 10 learners will distil the FL model with their own model. The result plots show the test accuracy averaged by testing a model on the ten individual parties ((\textcolor{blue}{\bf IND})). The x-axis shows for each approach the total number of local epochs the model was trained on the local dataset of the ten parties' devices.

\smartparagraph{Traditional AI Models: } first, we leverage a traditional logistic regression (LR) model to conduct experiments in a device heterogeneous setting. The LR model is trained on a non-IID synthetic dataset distributed over 10K clients.  As shown in \cref{fig:results-synthetic}, training for more epochs does not improve the model for the individual parties (\textcolor{blue}{\bf IND}). The FL model (\textcolor{ForestGreen}{\bf FL}) shows a better accuracy when tested on the ten parties. Therefore, the ten parties request the FL model via (\textcolor{red}{\bf MDD}) which when distilled to their local model (after only 5 local epochs) results in the best model with the highest accuracy.

\smartparagraph{Deep Neural Networks (DNN) Models:} next, we leverage DNN models such as Convolutional Neural Networks (CNN) and Recurrent Neural Networks (RNN) to conduct experiments in a device heterogeneous setting. The CNN and RNN models are trained on non-IID Femnist and Reddit datasets distributed over 3.4K and 813 clients, respectively.

\begin{figure}[!h]
    \centering
\includegraphics[width=1\columnwidth]{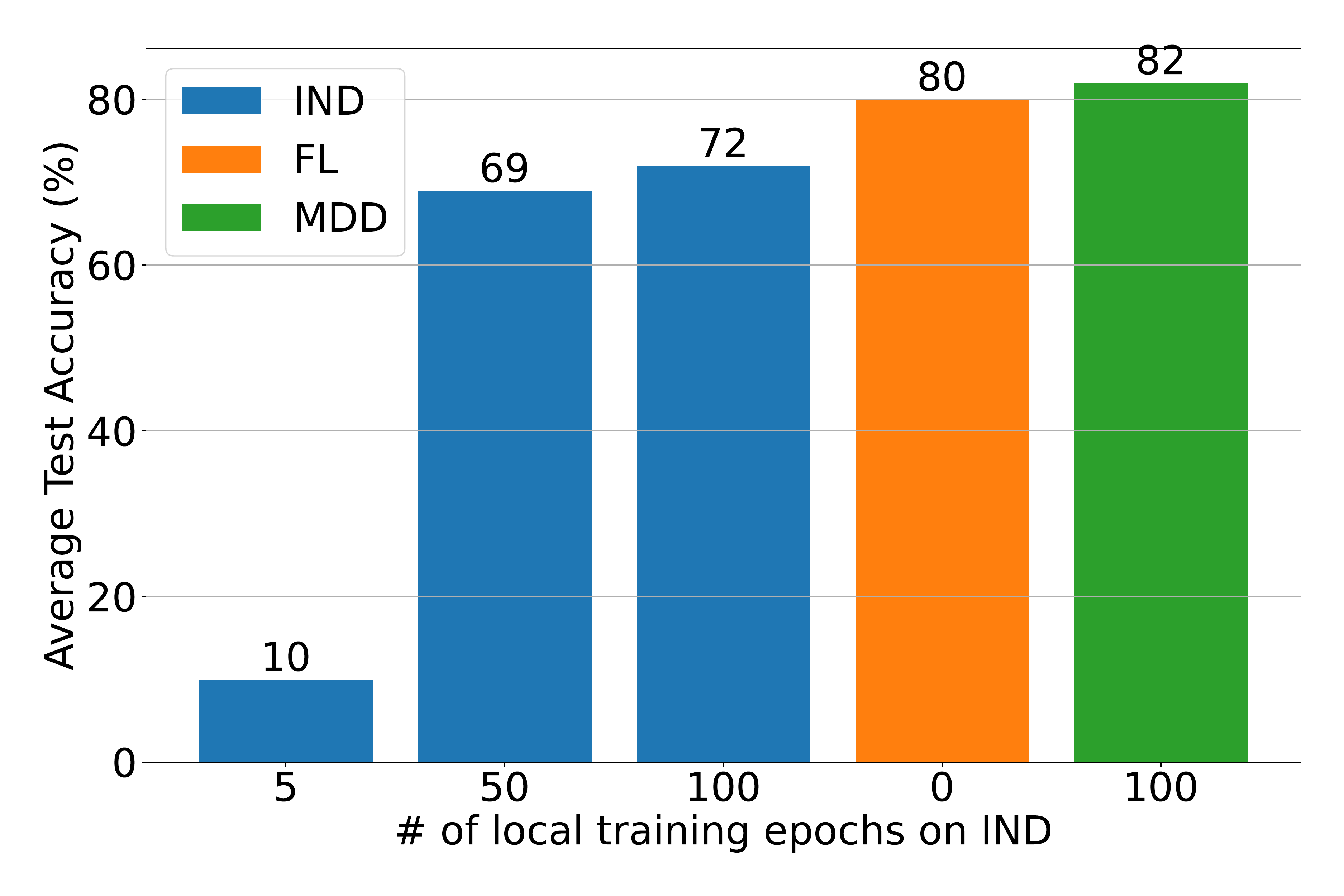}
    \caption{Performance comparison in the CNN-Femnist scenario.}
    \label{fig:results-femnist}
\end{figure}

Next, in \cref{fig:results-femnist}, we observe that training for the model locally for 100 epochs does improve the model for the individual parties (\textcolor{blue}{\bf IND}) and the FL model (\textcolor{ForestGreen}{\bf FL}) shows very low accuracy when tested on the ten parties. This might be due to the inability of the model to generalize well as the Femnist dataset which contains a large number of classes (i.e., digits and letters) with samples distributed over 3390 FL clients. We note then that if the individual parties (\textcolor{blue}{\bf IND}) request the FL model and distil their models with the FL model then the result is a better model in accuracy by more than 5 accuracy points (i.e., \textcolor{red}{\bf MDD}).

\begin{figure}[!h]
    \centering
\includegraphics[width=1\columnwidth]{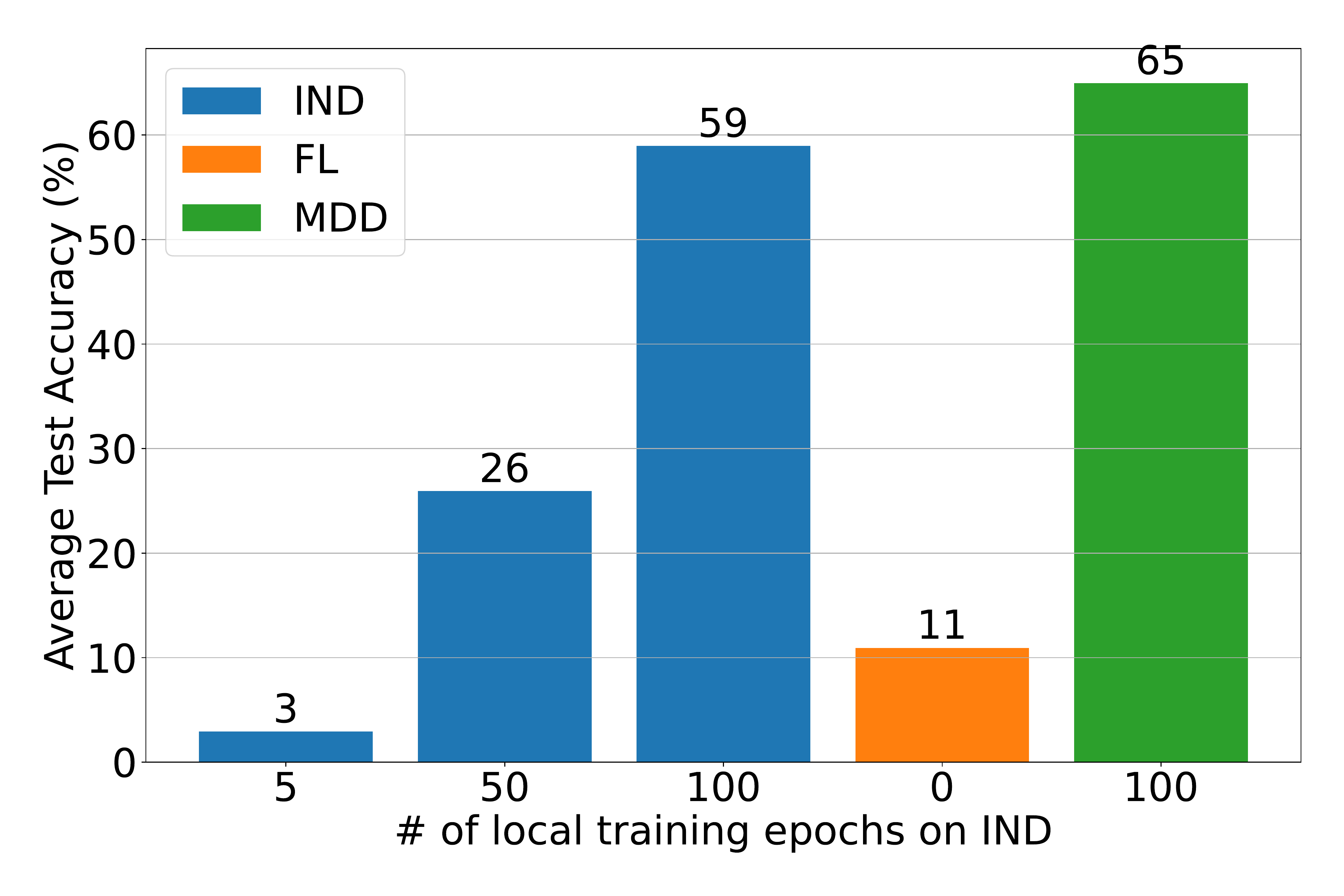}
    \caption{Performance comparison in the RNN-Reddit scenario.}
    \label{fig:results-reddit}
\end{figure}
  
Next, we observe in \cref{fig:results-reddit} that in the RNN-Reddit case training the model locally on the devices of the individual parties (\textcolor{blue}{\bf IND}) achieves comparable accuracy to the mode trained by the FL group (\textcolor{ForestGreen}{\bf FL}). This is because, in this scenario, the population are smaller and therefore the FL model could perform better on the \textcolor{blue}{\bf IND} group. Further, we observe a significantly better accuracy (nearly 16 accuracy points) is achieved by discovering and distilling the FL model with the model of the individual group (i.e., the \textcolor{red}{\bf MDD} approach).

In summary, the results, over various benchmarks and scenarios show the efficacy of the proposed MDD approach. However, this approach requires further investigation into several technical aspects to mature into practicality.

\section{Related Work}

\smartparagraph{Federated Learning (FL):} is a new distributed machine learning method that is becoming increasingly popular for its privacy-preserving and low-communication features. This motivated the growing adoption of FL to improve the end-user experience (e.g., the search suggestion quality of virtual keyboards~\cite{yang2018applied}). In FL, training a global model is assigned to a sub-population of decentralized devices such as mobile or IoT devices. These devices possess private data and engage in training the model on their  data~\cite{Bonawitz19,mcmahan2017}.   

\smartparagraph{System heterogeneity:} One of the major contributors to system performance unpredictability is the heterogeneity inherent in many distributed systems. Mainly, in the FL context, the heterogeneity of devices' system configurations (e.g., computation, communication, battery, etc) results in unpredictable performance. For instance, the stragglers (i.e., slow workers) can halt the training process for a prolonged duration~\cite{Ahmed-AQFL-21,Ahmed-REFL-2023}. 
Several solutions exist that address this problem
through system and algorithmic solutions~\cite{Li2020FedProx,Ahmed-AQFL-21,Ahmed-REFL-2023,Oort-osdi21}. Moreover, in FL, the heterogeneity is also a byproduct of other artifacts other than the devices. For example, the learner data distributions, the participants' selection method, and the behaviour of the device's owner are common sources of heterogeneity in FL setting~\cite{Ahmed-EuroMLSys-22,Ahmed-IoTJ-2023,Bonawitz19}. %

\smartparagraph{Energy-conservation:} Considering the uncertainties in the mobile environment, recently energy-aware federated learning techniques have been proposed~\cite{Amna-FedEdge-2022,kairouz2019advances}. These works aim for energy conservation to mitigate client dropouts which is the major contributor to the degraded FL model qualities~\cite{Amna-FedEdge-2022}.

\smartparagraph{Improvements in FL:} 
In FL, several works aim to improve the time-to-accuracy of training by leveraging techniques such as periodic updates, adaptive compression, and asynchronous updates~\cite{Bonawitz19,Ahmed-AQFL-21,Amna-FedEdge-2022,Ahmed-REFL-2023}.
Several methods that have been shown to achieve good performance in distributed ML settings could be applied to the FL settings such as~\cite{Ahmed-huffman-2020,grace-2021,Ahmed-SIDCo-MLSys21,SIDCOPATENT,Ahmed-DC2-INFOCOM21,Ahmed-NeurIPS-2021}
Other work studied the privacy guarantees of FL settings \cite{Melis2019,Nasr2019}. %
Moreover, the bias in FL is studied to ensure fair participation in the training  process~\cite{Ahmed-REFL-2023,Ahmed-ICC-2023}.

\section{Conclusion and Future Work}
This work discusses the difficulties that major decentralized learning paradigms face and demonstrates how they fall short of facilitating effective collaboration in complex decentralized environments. We emphasize the pressing requirement for the research community to put forth fresh, model-sharing-based designs. This encourages the services for identification of the models held by different parties in the network in order to facilitate large-scale, decentralized collaboration in learning. In the course of our ongoing research, we intend to create protocols that will enable scalable and secure management of model storage, discovery, and transfer within the suggested architecture.

\bibliographystyle{ieeetr}
\bibliography{refs,mypapers}

\end{document}